\documentclass{article}
\usepackage{spconf,amsmath,graphicx}
\usepackage{amssymb} 
\usepackage{caption}
\usepackage{subcaption}

\title{Pose Guided Person Image Generation with Hidden p-Norm Regression}
\name{Ting-Yao Hu, Alexander G. Hauptmann }
\address{Carnegie Mellon University, Pittsburgh, PA, USA}
\begin{document}
%
\maketitle
\begin{abstract}
In this paper, we propose a novel approach to solve the pose guided person image generation task.
We assume that the relation between pose and appearance information can be described by a simple matrix operation in hidden space.
Based on this assumption, our method estimates a pose-invariant feature matrix for each identity, and uses it to predict the target appearance conditioned on the  target pose.
The estimation process is formulated as a p-norm regression problem in hidden space.
By utilizing the differentiation of the solution of this regression problem, the parameters of the whole framework can be trained in an end-to-end manner.
While most previous works are only applicable to the supervised training and single-shot generation scenario, our method can be easily adapted to unsupervised training and multi-shot generation.
Extensive experiments on the challenging Market-1501 dataset show that our method yields competitive performance in all the aforementioned variant scenarios.

\end{abstract}

\section{Introduction}
\label{sec:intro}
In this paper, we deal with the task of generating realistic person images with the guidance of pose information.
Although Generative Adversarial Network (GAN) \cite{gan} allows computers to generate photo-realistic images, it is still difficult to capture the deformation from source pose to target pose.
This task has attracted the attention of researchers because it provides benefits to multiple applications, such as video synthesis \cite{liu2019liquid} and data augmentation for person re-identification \cite{qian2018reidaug}.

Several methods have been proposed to resolve the pose guided person image generation task \cite{ma2017pose,zhu2019progressive, tang2020xinggan, ren2020spatial, siarohin2018deformable}.
One type of approaches \cite{zhu2019progressive, tang2020xinggan} utilizes attention mechanisms to model the pose-appearance relation.
Another type of works \cite{ren2020spatial, siarohin2018deformable} relies on the deformation of hidden appearance feature map according to affine transformation.
While receiving promising performance, these methods usually lack flexibility comparing to real-world settings.
Specifically, they are developed based on the following two assumptions: (1) the availability of identity information of person images, and (2) the generation process is always conditioned on a single source image.
However, the first assumption is invalid if the human annotation resource is constrained, while the second becomes a limitation if we can collect multiple images of the same person during inference phase.
Although methods for unsupervised training \cite{song2019unsupervised} and multi-shot generation \cite{lathuiliere2020attention} have been investigated to resolve these two issues, respectively, they are still designed for a specific training/testing scenario.

In this paper, our goal is to develop a simple yet effective, flexible approach that is suitable for different situations: with/without identity information in training, single/multi-shot information in inference.
Hence, we propose a p-Norm regression (pNR) module, which models the relation among input appearance and pose feature matrices $H, P$, and a pose invariant feature set $F$ for each identity as a simple matrix operation in hidden space: $H \approx PF$.
Based on this design, pNR module estimates $F$ by solving a regression problem, and uses the optimal $F$ and the target pose feature $P_t$ to reconstruct the target appearance feature matrix $H_t$, which becomes the input of an image generator.
Then, comparing the generated image with ground truth target image, we can train the appearance/pose feature extractors, and the image generator in an end-to-end manner. 

In unsupervised training scenario, we use the pNR module to reconstruct the input appearance $H$ from partial observation of $H$, following the spirit of denoising auto-encoder \cite{vincent2008extracting}.
In multi-shot generation, we exploit multi-shot information to estimate pose-invariant feature matrix $F$ by constructing a larger regression problem.
These two strategies make our overall framework more flexible.

The main contributions of this paper are two fold:
(1) proposing p-Norm regression (pNR) module, which estimates pose-invariant feature and predicts the target appearance feature by solving a regression problem in hidden space.
(2) we demonstrate the applicability and efficacy of pNR module for pose guided person image generation task in supervised, unsupervised and multi-shot scenarios.


\section{Proposed Method}
\label{sec:method}

\begin{figure}[t]
    \centering
    \includegraphics[width=0.48\textwidth]{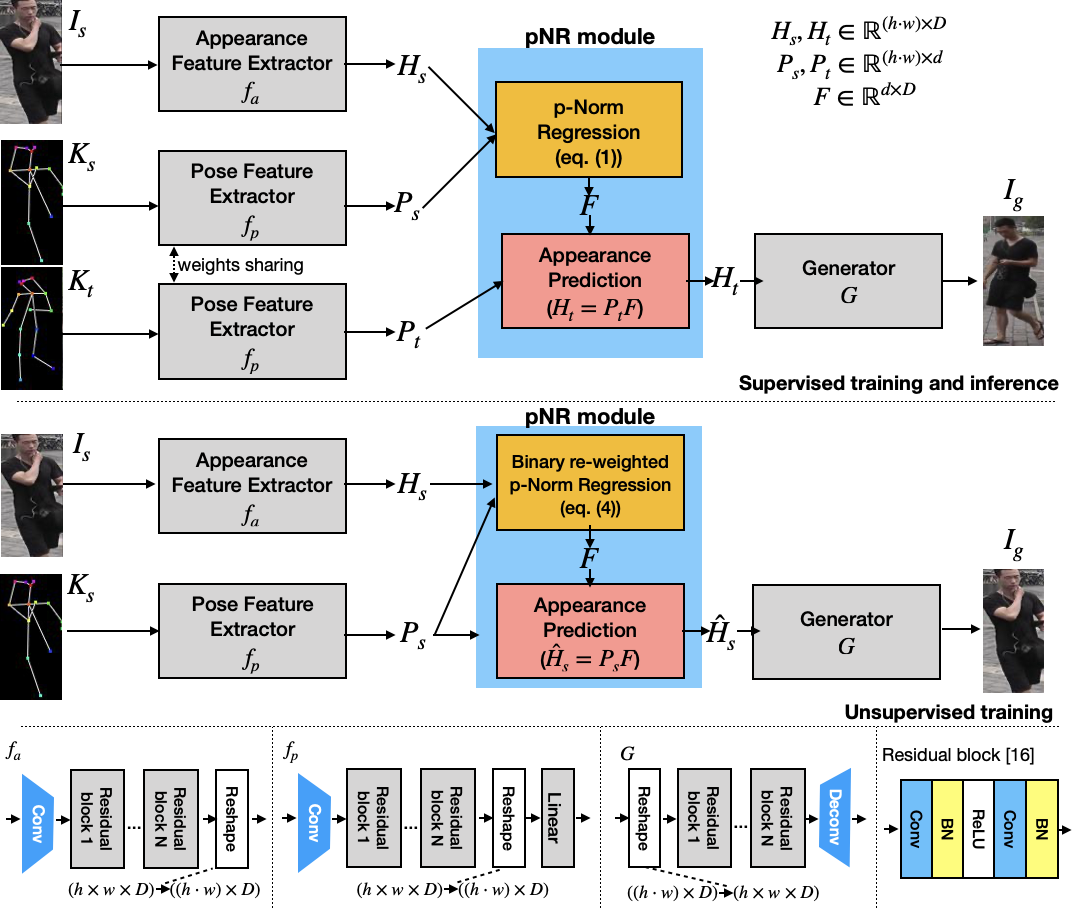}
    \caption{Overall architecture of our approach to pose guided person image generation. Our pNR module estimates a pose-invariant feature $F$ in hidden space, and exploits it to predict target appearance.}
    \label{fig:arch}
\end{figure}

We are given pairs of a person image and a key-point based representation of human pose, $(I, K)$.
Following previous works, $K$ is a keypoint heatmap of the 18 joints extracted from $I$ using Human Pose Estimator (HPE) \cite{cao2017realtime}.
Pose guided image generation aims at producing an image $I_g$ given a 3-tuple of source image, source pose, and target pose $(I_s, K_s, K_t)$.
The generated $I_g$ should reflect the pose described by $K_t$, and preserve the identity information of $I_s$, simultaneously.
The availability of the identity information is not always the same.
For example, when the human annotation resource is constrained, we may have no or little identity information in training phase.
Also, we may want to collect more information for a particular identity during inference phase, so the generation process would be conditioned on multiple pairs of $(I_s, K_s)$.
Hence, our ultimate goal is a effective and flexible approach, that is suitable for supervised/unsupervised training, and single/multi shot generation.
To achieve this goal, we propose a novel p-norm regression (pNR) module, along with a end-to-end learning framework.

\subsection{Overall system architecture}
Our system consists of the proposed pNR module, and three other components: pose and appearance feature extractors $f_a$, $f_p$, and an image generator $G$.
In the typical single-shot inference scenario, we are given a 3-tuple $(I_s, K_s, K_t)$.
Our system first extracts the appearance feature matrix $H_s \in \mathbb{R}^{ (h \cdot w) \times D }$ and pose feature matrices $P_s, P_t \in \mathbb{R}^{(h \cdot w) \times d} $
using appearance and pose feature extractors: $H_s = f_a(I_s), P_s = f_p(K_s), P_t = f_p(K_t)$.
Each row of $H$($P$) encodes the appearance(pose) characteristics at one of the $h \cdot w$ local regions in the raw image.
Then, our proposed pNR module estimates a pose-invariant feature $F \in \mathbb{R}^{d \times D}$ from $(H_s, P_s)$, and produces the target appearance feature matrix $H_t$.
Finally, the generator $G$ takes $H_t$ as input and produces the output image $I_g$.
In unsupervised scenario, identity information is unknown, so we can only exploit pairs of $(I_s, K_s)$ for training.
Thus, we solve a binary re-weighted version of p-norm regression, and leverage the reconstruction of $I_s$ as the supervision signal.
The overall architectures for all situations are illustrated in Fig. \ref{fig:arch}.

In the remaining of this section, we introduce the details of the proposed pNR module, and the strategies for unsupervised training and multi-shot generation.
Then, we elaborate the loss function for training.

\subsection{p-Norm regression (pNR) module}
Given appearance and pose feature matrices $H, P$ in hidden space, we assume that there exists a pose-invariant feature $F$, and three of them follows the simple relation: $H \approx PF$.
The motivation behind this assumption is that the appearance features at every spatial location share some common characteristics, which can be expressed by a set of $d$ feature vectors ($d \ll h \cdot w$).
Hence, each appearance feature (row of $H$) would be a linear combination of rows of $F$, and the combination weights are encoded in $P$,
Based on this motivation, the proposed p-norm regression (pNR) module contains two steps: estimation of $F$ and prediction of $H_t$.

In the first step, pNR module estimates $F$ from source appearance and pose information, $H_s, P_s$.
Specifically, the optimal $F$ is the solution of the following p-norm regression problem:
\begin{equation}
    F = \arg\min_{F'} ||H_s - P_s F'||_p
\label{eq:pnorm}
\end{equation}
In this work, we investigate two cases: $p=1$ and $p=2$.
If $p=2$, eq. (\ref{eq:pnorm}) would be a least square error (LSE) minimization problem, and $F$ can be calculated in closed form:
\begin{equation}
    F = (P_s^TP_s)^{-1} P_s^T H_s
\label{eq:lse}
\end{equation}
If $p=1$, eq. (\ref{eq:pnorm}) would become a least absolute deviation (LAD) problem, which is more robust to outliers. However, $F$ has no analytic solution.
In this case, we adopt iterative re-weighted least
square (IRLS) algorithm \cite{schlossmacher1973iterative}.
Let $F_t$ be the current estimation of $F$, the update rule for $F_{t+1}$ in the next iteration can be expressed as: 
\begin{equation}
    F_{t+1}^{(i)} = (P_s^T W_t^{(i)} P_s)^{-1} P_s^T W_t^{(i)} H_s^{(i)}
\label{eq:irls_f}
\end{equation}
where $F_{t+1}^{(i)}$ and $H_s^{(i)}$ are the $i$-th column of $F_{t+1}$ and $i$-th column of $H_s$, respectively, $i=1,...,D$.
$W_t^{(i)}$ is a $(h \cdot w) \times (h \cdot w)$ diagonal matrix, whose diagonal elements are from the $(h \cdot w)$-dimensional vector, $1/|H_s^{(i)} - P_s F_t^{(i)}|$.
In our implementation, we use the solution of LSE as initial $F_0$ and execute this update rule for a fix number of iteration.
The result of the last iteration would be assigned to $F$.
Given $F$, the second step of pNR module predicts target appearance based on the same assumption, $H_t = P_t F$.

Since pNR module is treated as an intermediate layer of the whole network architecture, we need to differentiate through it in order to train the parameters in $f_a, f_p$ and $G$ with SGD-like algorithm.
In the case of LSE ($p=2$), the derivative of $F$ in eq. (\ref{eq:lse}) can be obtained easily.
However, when $p=1$, the calculation of $F$ is an iterative process, so the precise derivative is difficult to compute.
Considering eq. (\ref{eq:irls_f}) in the last iteration of IRLS update, we calculate $\partial F^{(i)}/ \partial H_s^{(i)}, \partial F^{(i)}/ \partial P_s$ only, and 
ignore the derivative with respective to the recursive term, $\partial F^{(i)}/ \partial W_t^{(i)}$, during the backward propagation.
Although this calculation is an approximation, it still aims to preserve the robustness of LAD to outliers, and receives good empirical performance.

\subsection{Unsupervised training and multi-shot generation}

In unsupervised training, pNR module estimates $F$ by solving a binary re-weighted p-norm regression:
\begin{equation}
    F = \arg\min_{F'}  \sum_{j=1}^{h \cdot w} v_i||H_s^{j} - P_s^{j} F'||_p
\label{eq:unsup}
\end{equation}
where $v_j \in \{0,1\}$ is a binary random variable,
which blocks out the information from the $j$-th row in $H_s$ ($H_s^j$) when $v_j=0$.
In our case, we set $p(v_i=0) = p(v_i=1) = 0.5$.
Then, we predict the source appearance feature $\hat{H_s}=P_s F$ in the second step, and use $\hat{H_s}$ to reconstruct the input source image.

In multi-shot generation, we aggregate the information from $M$ pairs of source image and pose map $(I_s, K_s)$.
Specifically, we concatenate all the appearance and pose feature matrices, and construct a larger p-norm regression problem. 
It can still be expressed by eq. (\ref{eq:pnorm}), but $H_s \in \mathbb{R}^{ ( M \cdot h \cdot w) \times D }, P_s \in \mathbb{R}^{ (M \cdot h \cdot w) \times d }$.
Please note that $M$ can be different in training and inference phase, and can also be a varied number.

\subsection{Loss functions}
Following the design of previous works \cite{zhu2019progressive, tang2020xinggan}, the training objective of our system framework contains four components: L1 loss, Perceptual loss, $GAN_I$ loss, $GAN_K$ loss.

L1 loss computes the pixel-wise L1 distance between generated image and target image: $\mathcal{L}_{L1} = ||I_t - I_g||_1$.
Perceptual loss compares two images in the space of pretrained features: $\mathcal{L}_{per} =  ||\Phi_{\rho}(I_t) - \Phi_{\rho}(I_g)||_1$,
where $\Phi$ is a VGG19 \cite{simonyan2014very} network pretrained on ImageNet \cite{russakovsky2015imagenet}, and $\rho$ is the index of hidden layers ($\rho = Conv1 \_ 2$ in our case). On the other hand, the purpose of $GAN_I$ and $GAN_K$ loss is to align the output of generator to two probability distributions: $p(I_t|I_s)$ and $p(I_t|K_t)$, respectively.
To do so, we measure the two types of distribution discrepancy by two discriminators, $D_I$ and $D_K$, respectively.
The former distinguishes generated images from target images conditioned on source image $I_s$, while the later does so conditioned on the pose map $K_t$.
Thus, two loss functions are formulated as:
\begin{equation}
\begin{split}
    \mathcal{L}_{GAN_I} =& E[\log (D_I(I_t, I_s))] + E [\log (1-D_I(I_g, I_s))] \\
    \mathcal{L}_{GAN_K} =& E[\log (D_K(I_t,K_t))]+ E[\log (1-D_K(I_g,K_t))]
\end{split}
\end{equation}
where the expectation is computed over the distribution of $I_s, I_t$ pairs.
The overall training objective is the weighted combination of the four components, and the training process can be expressed as:
\begin{equation}
  \min_{f_a, f_p, G}\;\max_{D_I,D_K} \lambda_1 \mathcal{L}_{L1} + \lambda_2 \mathcal{L}_{per} + \lambda_3 \mathcal{L}_{GAN_I} + \lambda_4 \mathcal{L}_{GAN_K}
\end{equation}
In unsupervised training scenario, source and target images are the same, so we replace $(I_t, K_t)$ to $(I_s, K_s)$ in all the loss functions, and disable $\mathcal{L}_{GAN_I}$ by setting $\lambda_3 = 0$.

\section{Experiment}
\label{sec:ep}
\begin{table}[t]
\begin{tabular}{l|llll}
\hline
                                   & IS             & SSIM           & mask-IS        & mask-SSIM      \\ \hline
PG$^2$ \cite{ma2017pose}             & 3.460          & 0.253          & 3.435          & 0.792          \\
Def-GAN \cite{siarohin2018deformable}                             & 3.185          & 0.290          & 3.502          & 0.805          \\
PATN    \cite{zhu2019progressive}            & 3.323          & 0.311          & 3.773          & 0.811    \\
XingGAN \cite{tang2020xinggan}                           & 3.506          & \textbf{0.313} & \textbf{3.872} & \textbf{0.816} \\
pNR (LSE)                    & 3.435          & 0.298          & 3.741          & 0.802          \\
pNR (LAD)        & \textbf{3.631}          & 0.305          & 3.796          & 0.807          \\ \hline\hline
SPT$^*$  \cite{song2019unsupervised}     & 3.449          & 0.203          & \textbf{3.680} & 0.758          \\
pNR$^*$ (LSE) & \textbf{3.688} & 0.241          & 3.501          & 0.783          \\
pNR$^*$ (LAD) & 3.681          & \textbf{0.248} & 3.610          & \textbf{0.789} \\ \hline
\end{tabular}
\caption{Quantitative results on Market-1501. All the metrics are the higher the better. *: Unsupervised training.}
\label{tab:market}
\end{table}

\begin{figure}
    \centering
    \includegraphics[width=0.45\textwidth]{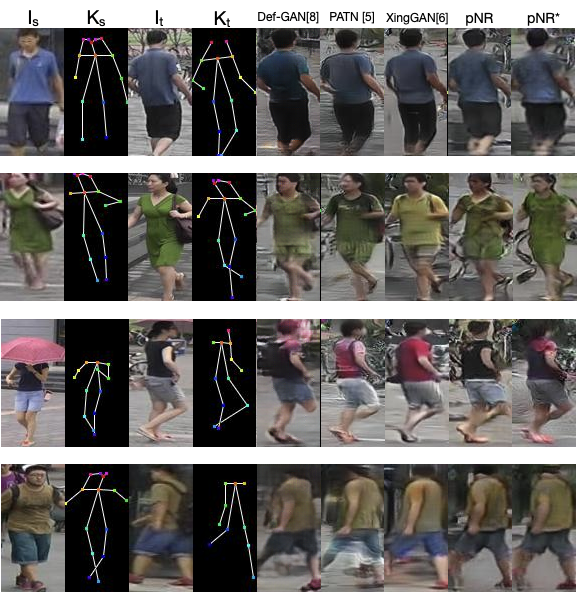}
    \caption{Qualitative comparison between pNR module and other methods. *: Unsupervised training.}
    \label{fig:qualitative}
\end{figure}

\subsection{Implementation}
Pose and appearance feature extractors, $f_p$ and $f_a$ both consist of two downsampling CNN layers, a sequence of N residual blocks proposed in \cite{johnson2016perceptual} and a reshape operator, while $f_p$ has another linear layer to reduce the dimension to $d$.
Image generator $G$ contains N residual blocks followed by two unsampling deconv layers.
We set $N=4$ and $N=2$ in supervised and unsupervised training, respectively.
For the discriminators $D_I$ and $D_K$, we exploit the same implementation as in \cite{zhu2019progressive}.
For pNR module, we set $d=20, D=256$.

In the training phase, Adam \cite{kingma2014adam} optimizer with learning rate of 0.002, $\beta_1 = 0.5$, $\beta_2 = 0.999$ is adopted.
For the hyper-parameters of loss function, we set $\lambda_1 = \lambda_2 = 5, \lambda_3 = \lambda_4 = 10$ for supervised training, and change $\lambda_3$ to 0 for unsupervised training.

\subsection{Dataset and evaluation protocols}
We evaluate our pNR module on the challenging Market-1501 \cite{zheng2015scalable} dataset, which was designed for person re-identification.
Performing pose guided image generation on this dataset is challenging because of its low resolution ($128 \times 64$), and high diversity in pose, background and illumination.
Following previous works, we detect the keypoint-based pose representation by HPE, and remove images in which no human body can be detected.
Consequently, the training and single-shot testing sets consists of 263,632 and 12,000 pairs of images with the same identity.
Sets of identities for training and testing are mutually exclusive.
For multi-shot generation, we keep those identities in single-shot testing set with 6 or more images, and sample 12,000 tuples with 5 source images and 1 target image for testing.

In all of the experiments, we adopt Structure Similarity (SSIM) \cite{wang2004image} and Inception Score (IS) \cite{is_nips} as the evaluation metrics.
SSIM measures correctness of pose transfer by comparing generated and ground truth images, while IS uses a pretrained image classifier to assess the image quality.
The masked version of both metrics are also utilized to reduce the distraction from irrelevant background regions.

\subsection{Comparison with state-of-the-art}
We compare our proposed method with some previous state-of-the-art, including PG$^2$ \cite{ma2017pose}, Def-GAN \cite{siarohin2018deformable}, PATN \cite{zhu2019progressive}, XingGAN \cite{tang2020xinggan}, and SPT \cite{song2019unsupervised} (unsupervised). 
From the results shown in Table \ref{tab:market}, we make the following observations.
First, pNR with LAD ($p=1$) performs better than that with LSE ($p=2$) in both supervised and unsupervised training scenarios.
Second, in supervised training, pNR yields competitive performance compared with most recent state-of-the-art, PATN and XingGAN.
Third, in unsupervised training, pNR$^*$ outperforms the previous work, SPT, by a large margin on every metrics except mask-IS.

We present a qualitative study in Fig. \ref{fig:qualitative}.
Compared with other previous works, our method is more capable of capturing pose and appearance information.
We also list the results from pNR$^*$ (unsupervised training), which are still in good quality but contain more artifacts than supervised methods.

\subsection{Multi-shot generation}
\begin{figure}
    \centering
    \includegraphics[width=0.45\textwidth]{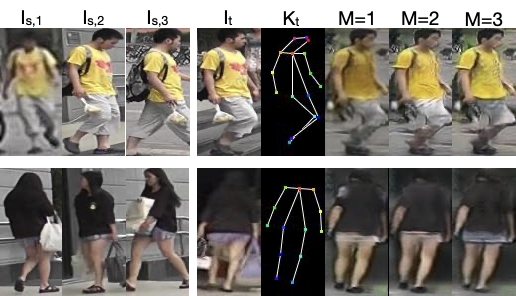}
    \caption{Qualitative results of multi-shot generation using pNR module (supervised training, LAD).}
    \label{fig:multishot}
\end{figure}

This experiment demonstrates that our method can effectively integrate information from multi-shot source images.
For model training, we apply LAD ($p=1$) in pNR module, and exploit single-shot ($M=1$) training dataset.
From the quantitative comparison in Table \ref{tab:multi}, we can see that our pNR module outperforms previous work \cite{lathuiliere2020attention}, and
yields better SSIM and mask-SSIM with larger number of source images in both supervised and unsupervised training scenarios.
Also, the qualitative results in Fig. \ref{fig:multishot} show that pNR module reconstructs more details when more source images are available.

\begin{table}[]
\centering
\begin{tabular}{l|llll}
\hline
              & IS    & SSIM  & mask-IS & mask-SSIM \\ \hline
\cite{lathuiliere2020attention}, M=1 & 3.251 & 0.270 & 3.614   & 0.771     \\
\cite{lathuiliere2020attention}, M=3 & 3.442 & 0.291 & 3.739   & 0.783     \\
\cite{lathuiliere2020attention}, M=5 & 3.444 & 0.306 & \textbf{3.814}   & 0.788     \\ \hline
pNR, M=1      & 3.631 & 0.305 & 3.796   & 0.807     \\
pNR, M=3      & 3.642 & 0.311 & 3.804   & 0.812     \\
pNR, M=5      & 3.629 & \textbf{0.313} & 3.804   & \textbf{0.818}     \\ \hline
pNR*, M=1     & \textbf{3.684} & 0.248 & 3.610   & 0.789     \\
pNR*, M=3     & 3.640 & 0.254 & 3.616   & 0.801     \\
pNR*, M=5     & 3.662 & 0.259 & 3.614   & 0.805     \\ \hline
\end{tabular}
\caption{Results of multi-shot generation. All the metrics are the higher the better. *: Unsupervised training}
\label{tab:multi}
\end{table}

\section{Conclusion}
\label{sec:conclusion}
We propose a novel pNR module to tackle pose guided person image generation task.
It estimates a pose-invariant feature matrix for each identity and predicts the target appearance feature by solving a p-norm regression problem in hidden space.
Integrated with CNN-based pose/appearance feature extractors, pNR module serves as a layer of the whole network architecture and supports end-to-end training.
The experiment results demonstrate the efficacy of the pNR module in a supervised and unsupervised training scenario, as well as generating images from multi-shot person image data.


\bibliographystyle{IEEE}
\bibliography{strings,refs}

\begin{thebibliography}{10}

\bibitem{gan}
Ian Goodfellow, Jean Pouget-Abadie, Mehdi Mirza, Bing Xu, David Warde-Farley,
  Sherjil Ozair, Aaron Courville, and Yoshua Bengio,
\newblock ``Generative adversarial nets,''
\newblock in {\em Advances in Neural Information Processing Systems},
  Z.~Ghahramani, M.~Welling, C.~Cortes, N.~Lawrence, and K.~Q. Weinberger,
  Eds., 2014, pp. 2672--2680.

\bibitem{liu2019liquid}
Wen Liu, Zhixin Piao, Jie Min, Wenhan Luo, Lin Ma, and Shenghua Gao,
\newblock ``Liquid warping gan: A unified framework for human motion imitation,
  appearance transfer and novel view synthesis,''
\newblock in {\em Proceedings of the IEEE International Conference on Computer
  Vision}, 2019, pp. 5904--5913.

\bibitem{qian2018reidaug}
Xuelin Qian, Yanwei Fu, Tao Xiang, Wenxuan Wang, Jie Qiu, Yang Wu, Yu-Gang
  Jiang, and Xiangyang Xue,
\newblock ``Pose-normalized image generation for person re-identification,''
\newblock in {\em Proceedings of the European conference on computer vision
  (ECCV)}, 2018, pp. 650--667.

\bibitem{ma2017pose}
Liqian Ma, Xu~Jia, Qianru Sun, Bernt Schiele, Tinne Tuytelaars, and Luc
  Van~Gool,
\newblock ``Pose guided person image generation,''
\newblock in {\em Advances in neural information processing systems}, 2017, pp.
  406--416.

\bibitem{zhu2019progressive}
Zhen Zhu, Tengteng Huang, Baoguang Shi, Miao Yu, Bofei Wang, and Xiang Bai,
\newblock ``Progressive pose attention transfer for person image generation,''
\newblock in {\em Proceedings of the IEEE Conference on Computer Vision and
  Pattern Recognition}, 2019, pp. 2347--2356.

\bibitem{tang2020xinggan}
Hao Tang, Song Bai, Li~Zhang, Philip~HS Torr, and Nicu Sebe,
\newblock ``Xinggan for person image generation,''
\newblock in {\em European Conference on Computer Vision}. Springer, 2020, pp.
  717--734.

\bibitem{ren2020spatial}
Yurui Ren, Xiaoming Yu, Junming Chen, Thomas~H Li, and Ge~Li,
\newblock ``Deep image spatial transformation for person image generation,''
\newblock in {\em Proceedings of the IEEE/CVF Conference on Computer Vision and
  Pattern Recognition}, 2020, pp. 7690--7699.

\bibitem{siarohin2018deformable}
Aliaksandr Siarohin, Enver Sangineto, St{\'e}phane Lathuiliere, and Nicu Sebe,
\newblock ``Deformable gans for pose-based human image generation,''
\newblock in {\em Proceedings of the IEEE Conference on Computer Vision and
  Pattern Recognition}, 2018, pp. 3408--3416.

\bibitem{song2019unsupervised}
Sijie Song, Wei Zhang, Jiaying Liu, and Tao Mei,
\newblock ``Unsupervised person image generation with semantic parsing
  transformation,''
\newblock in {\em Proceedings of the IEEE Conference on Computer Vision and
  Pattern Recognition}, 2019, pp. 2357--2366.

\bibitem{lathuiliere2020attention}
St{\'e}phane Lathuili{\`e}re, Enver Sangineto, Aliaksandr Siarohin, and Nicu
  Sebe,
\newblock ``Attention-based fusion for multi-source human image generation,''
\newblock in {\em The IEEE Winter Conference on Applications of Computer
  Vision}, 2020, pp. 439--448.

\bibitem{vincent2008extracting}
Pascal Vincent, Hugo Larochelle, Yoshua Bengio, and Pierre-Antoine Manzagol,
\newblock ``Extracting and composing robust features with denoising
  autoencoders,''
\newblock in {\em Proceedings of the 25th international conference on Machine
  learning}, 2008, pp. 1096--1103.

\bibitem{cao2017realtime}
Zhe Cao, Tomas Simon, Shih-En Wei, and Yaser Sheikh,
\newblock ``Realtime multi-person 2d pose estimation using part affinity
  fields,''
\newblock in {\em Proceedings of the IEEE conference on computer vision and
  pattern recognition}, 2017, pp. 7291--7299.

\bibitem{schlossmacher1973iterative}
EJ~Schlossmacher,
\newblock ``An iterative technique for absolute deviations curve fitting,''
\newblock {\em Journal of the American Statistical Association}, 1973.

\bibitem{simonyan2014very}
Karen Simonyan and Andrew Zisserman,
\newblock ``Very deep convolutional networks for large-scale image
  recognition,''
\newblock {\em arXiv preprint arXiv:1409.1556}, 2014.

\bibitem{russakovsky2015imagenet}
Olga Russakovsky, Jia Deng, Hao Su, Jonathan Krause, Sanjeev Satheesh, Sean Ma,
  Zhiheng Huang, Andrej Karpathy, Aditya Khosla, Michael Bernstein, et~al.,
\newblock ``Imagenet large scale visual recognition challenge,''
\newblock {\em International journal of computer vision}, 2015.

\bibitem{johnson2016perceptual}
Justin Johnson, Alexandre Alahi, and Li~Fei-Fei,
\newblock ``Perceptual losses for real-time style transfer and
  super-resolution,''
\newblock in {\em European conference on computer vision}. Springer, 2016, pp.
  694--711.

\bibitem{kingma2014adam}
Diederik~P Kingma and Jimmy Ba,
\newblock ``Adam: A method for stochastic optimization,''
\newblock {\em arXiv preprint arXiv:1412.6980}, 2014.

\bibitem{zheng2015scalable}
Liang Zheng, Liyue Shen, Lu~Tian, Shengjin Wang, Jingdong Wang, and Qi~Tian,
\newblock ``Scalable person re-identification: A benchmark,''
\newblock in {\em Proceedings of the IEEE international conference on computer
  vision}, 2015.

\bibitem{wang2004image}
Zhou Wang, Alan~C Bovik, Hamid~R Sheikh, and Eero~P Simoncelli,
\newblock ``Image quality assessment: from error visibility to structural
  similarity,''
\newblock {\em IEEE transactions on image processing}, vol. 13, no. 4, pp.
  600--612, 2004.

\bibitem{is_nips}
Tim Salimans, Ian Goodfellow, Wojciech Zaremba, Vicki Cheung, Alec Radford,
  Xi~Chen, and Xi~Chen,
\newblock ``Improved techniques for training gans,''
\newblock in {\em Advances in Neural Information Processing Systems}, D.~Lee,
  M.~Sugiyama, U.~Luxburg, I.~Guyon, and R.~Garnett, Eds., 2016, pp.
  2234--2242.

\end{thebibliography}

\end{document}